\documentclass[runningheads]{llncs}
\usepackage{graphicx}
\usepackage{amsmath,amssymb} 
\usepackage{color}
\usepackage{times}
\usepackage{epsfig}

\usepackage{enumitem}
\usepackage{booktabs}
\usepackage{subfigure}
\usepackage{multirow}
\usepackage{mathtools}
\usepackage{soul}
\usepackage{algorithm}
\usepackage{algorithmic}
\usepackage{bbm}

\newcommand{\Paragraph}[1]{\vspace{1mm} \noindent \textbf{#1} \hspace{0mm}}

\newcommand{\model}{CoSCA}

\begin{document}
\pagestyle{headings}
\mainmatter

\def\ACCV20SubNumber{721}  

\title{Contrastively Smoothed Class Alignment for Unsupervised Domain Adaptation} 
\titlerunning{CoSCA for Unsupervised Domain Adaptation}
%
\author{Shuyang Dai\inst{1} \and
Yu Cheng\inst{2} \and
Yizhe Zhang\inst{3} \and
Zhe Gan\inst{2} \and \\
Jingjing Liu\inst{2} \and
Lawrence Carin\inst{1}}
\authorrunning{S. Dai et al.}
%
\institute{$^1$~Duke University,~$^2$~Microsoft Dynamics 365 AI Research,~$^3$~Microsoft Research
\email{\{shuyang.dai,lcarin\}@duke.edu}
\email{\{yu.cheng,yizhe.zhang,zhe.gan,jingjl\}@microsoft.com}}

\maketitle

\begin{abstract}
Recent unsupervised approaches to domain adaptation primarily focus on minimizing the gap between the source and the target domains through refining the feature generator, in order to learn a better alignment between the two domains. This minimization can be achieved via a domain classifier to detect target-domain features that are divergent from source-domain features. 
However, when optimizing via such domain-classification discrepancy, ambiguous target samples that are not smoothly distributed on the low-dimensional data manifold are often missed.
To solve this issue, we propose a novel Contrastively Smoothed Class Alignment (\model) model, that explicitly incorporates both intra- and inter-class domain discrepancy to better align ambiguous target samples with the source domain. \model~estimates the underlying label hypothesis of target samples, and simultaneously adapts their feature representations by optimizing a proposed contrastive loss. In addition, Maximum Mean Discrepancy (MMD) is utilized to directly match features between source and target samples for better global alignment. 
Experiments on several benchmark datasets demonstrate that \model outperforms state-of-the-art approaches for unsupervised domain adaptation by producing more discriminative features.
\end{abstract}

\section{Introduction}
Deep neural networks (DNNs) have significantly improved the state of the art on many supervised tasks~\cite{donahue2014decaf,yosinski2014transferable,simonyan2014very,he2016deep}. However, without sufficient training data, DNNs often generalize poorly to new tasks or new environments~\cite{torralba2011unbiased}. This is known as dataset bias or a domain-shift problem \cite{gretton2009covariate}.
Unsupervised domain adaptation (UDA) \cite{pan2010survey,ganin2016domain} aims to generalize a model learned from a source domain with rich annotated data to a new target domain without any labeled data. Recently, many approaches have been proposed to learn transferable representations, by simultaneously matching feature distributions across different domains \cite{haeusser2017associative,tzeng2015simultaneous}. 

\begin{figure}[!t]
    \centering
    \includegraphics[width=0.8\textwidth]{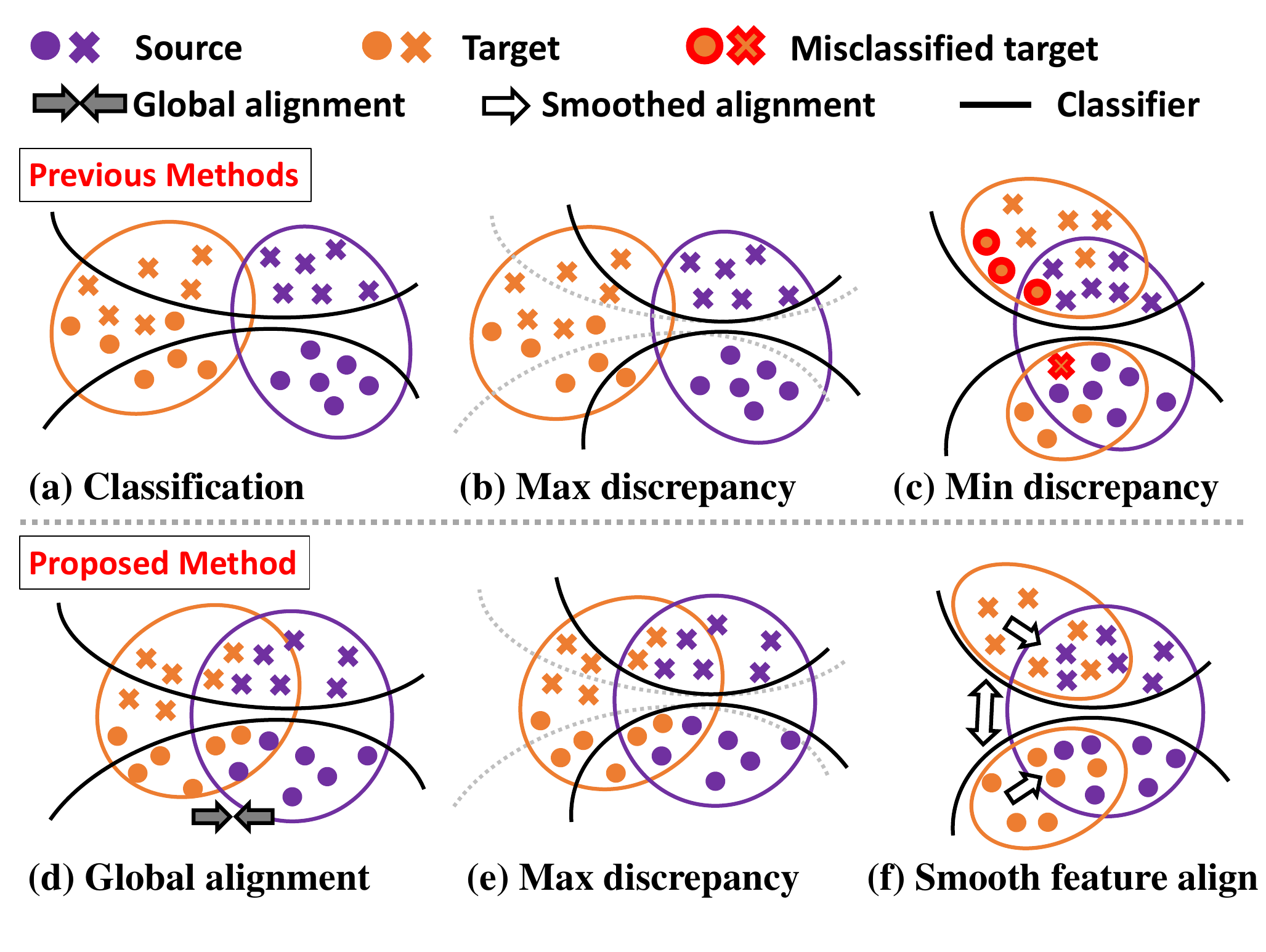}
    \caption{Comparison between previous classifier-discrepancy-based methods and our proposed \model~in the \textit{feature} space. \textbf{Top:} The region of vacancy created by maximum discrepancy reduces the smoothness of alignment between ambiguous target samples and source samples, leading to sub-optimal solutions. This problem becomes more severe when global domain alignment is not considered. \textbf{Bottom:} Demonstration of global alignment and class-conditional adaptation by using the proposed \model. After classifier discrepancy is maximized, the proposed contrastive loss moves ambiguous target samples near the decision boundary towards their neighbors and separates them from non-neighbors.}
    \label{fig:intuition}
\end{figure}

Motivated by~\cite{goodfellow2014generative}, \cite{tzeng2017adversarial,ganin2014unsupervised} introduced a min-max game: a domain discriminator is learned by minimizing the error of distinguishing data samples from the source and the target domains,
while a feature generator learns transferable features that are indistinguishable by the domain discriminator. This imposes that the learned features are domain-invariant. Additionally, a feature classifier ensures that the learned features are discriminative in the source domain. Despite promising results, these adversarial methods suffer from inherent algorithmic weaknesses~\cite{shu2018dirt}. Specifically, the generator may manifest ambiguous features near class boundaries~\cite{saito2017maximum}: while the generator manages to fool the discriminator, some target-domain features may still be misclassified.
In other words, the model merely aligns the global marginal distribution of the two domains and ignores the class-conditional decision boundaries.

To overcome this issue, recent UDA models further align class-level distributions by taking the decision boundary into consideration. These methods either rely on iteratively refining the decision boundary with empirical data \cite{shu2018dirt,DBLP:journals/corr/abs-1711-01575}, or utilizing multi-view information \cite{kumar2018co}. Alternatively, the maximum classifier discrepancy (MCD)~\cite{saito2017maximum} model 
conducts a min-max game between a feature generator and two classifiers. Ambiguous target samples that are far from source-domain samples can be detected when the discrepancy between the two classifiers is maximized, as shown in Figure~\ref{fig:intuition}(b). Meanwhile, as the generator fools the classifiers, the generated target features may fall into the source feature regions.
However, the target samples may not be smooth on the low-dimensional manifold \cite{chapelle2009semi,luo2017smooth}, meaning that neighboring samples may not belong to the same class. As a result, some generated target features could be miscategorized as shown in Figure~\ref{fig:intuition}(c). 

In this paper, we propose the \textbf{Co}ntrastively \textbf{S}moothed \textbf{C}lass \textbf{A}lignment (\model) model to improve the latent alignment of class-conditional feature distributions between source and target domains, by alternatively estimating the underlying label hypothesis of target samples to map them into tighter clusters, and adapt feature representations based on a proposed contrastive loss. 
Specifically, by aligning ambiguous target samples near the decision boundaries with their neighbors and distancing them from non-neighbors, \model~enhances the alignment of each class in a contrastive manner. Figure~\ref{fig:intuition}(f) demonstrates an enhanced and smoothed version of the class-conditional alignment. 
Moreover, as shown in Figure~\ref{fig:intuition}(d), Maximum Mean Discrepancy (MMD) is included to better merge the source and target domain feature representations. The overall framework is trained end-to-end in an adversarial manner.

Our principal contributions are summarized as follows:
\begin{itemize}
        \item We propose CoSCA, a novel approach that smooths class alignment for maximizing classifier discrepancy with a contrastive loss.
        CoSCA also provides better global domain alignment via the use of MMD loss. 
        \item We validate the proposed approach on several domain adaptation benchmarks. Extensive experiments demonstrate that \model~achieves state-of-the-art results on several benchmarks. 
\end{itemize}

\section{Related Work}
\Paragraph{Unsupervised Domain Adaptation}
A practical solution for domain adaptation is to learn domain-invariant features whose distribution is similar across the source and the target domains. For example, Sener \textit{et al.} \cite{NIPS2016_6360} proposed using clustering techniques and pseudo-labels to obtain discriminative features. Long \textit{et al.} proposed DAN \cite{long2015learning} and JAN \cite{long2016unsupervised} to minimize the MMD or variations of MMD between two domains. Adversarial domain adaptation integrates adversarial learning and domain adaptation in a two-player game \cite{ganin2014unsupervised,tzeng2017adversarial,tzeng2015simultaneous}. Following this idea, most existing adversarial learning methods reduce feature differences by fooling a domain discriminator \cite{NIPS2018_7436,ganin2016domain}. However, these methods fail to consider the relationship between target samples and the class-conditional decision boundaries when aligning features~\cite{saito2017maximum}, while only merging the source and the target domains. 

\Paragraph{Class-conditional Alignment}
To address the aforementioned issue, recent work enforces class-level alignment while aligning global marginal distributions. Associative domain adaptation (ADA) \cite{haeusser2017associative} reinforces associations across domains directly in the embedding space, to extract features that are statistically domain-invariant and class-discriminative. 
Adversarial Dropout Regularization (ADR) \cite{DBLP:journals/corr/abs-1711-01575} and Maximum Classifier Discrepancy (MCD) \cite{saito2017maximum} were proposed to train a neural network in an adversarial manner, avoiding generating non-discriminative features lying in the region near the decision boundary. In \cite{DBLP:journals/corr/abs-1809-02176,long2016unsupervised} 
the authors considered class information when measuring domain discrepancy. Co-regularized Domain Adaptation (Co-DA)~\cite{kumar2018co} utilized multi-view information to match the marginal feature distributions corresponding to the class-conditional distributions. Compared with previous work that executed the alignment by optimizing ``hard" metrics \cite{saito2017maximum,kumar2018co}, we propose to smooth the alignment iteratively, with explicitly defined loss. 

\Paragraph{Contrastive Learning}
The intuition for contrastive learning is to let the model understand the difference between one set ($e.g.$, data points) and another, instead of only characterizing a single set \cite{NIPS2013_5007}. This idea has been explored in previous works that 
model intra-class compactness and inter-class
separability ($e.g.$, distinctiveness loss \cite{NIPS2017_6691},
contrastive loss \cite{Hadsell:2006:DRL:1153171.1153654}, triplet loss \cite{tripletrank}) and tangent distance \cite{tangent}.
It has also been extended to consider several assumptions in semi-supervised and unsupervised learning \cite{luo2017smooth,NIPS2017_6997}, such as the low-density region (or cluster) assumption \cite{luo2017smooth,tangent} that the decision boundary should lie in the low-density region, rather than crossing the high-density region. 
Recently, contrastive learning was applied in UDA \cite{kang2019contrastive}, in which the intra/inter-class domain discrepancy were modeled.
In comparison, our work is based on the MCD framework, utilizing the low-density assumption and focusing on separating the ambiguous target data points by optimizing the contrastive objective, allowing the decision boundary to sit in the low-density region, $i.e.$, region of vacancy, and smoothness assumption. 

\begin{figure*}[t]
    \centering
    \includegraphics[width=0.98\textwidth]{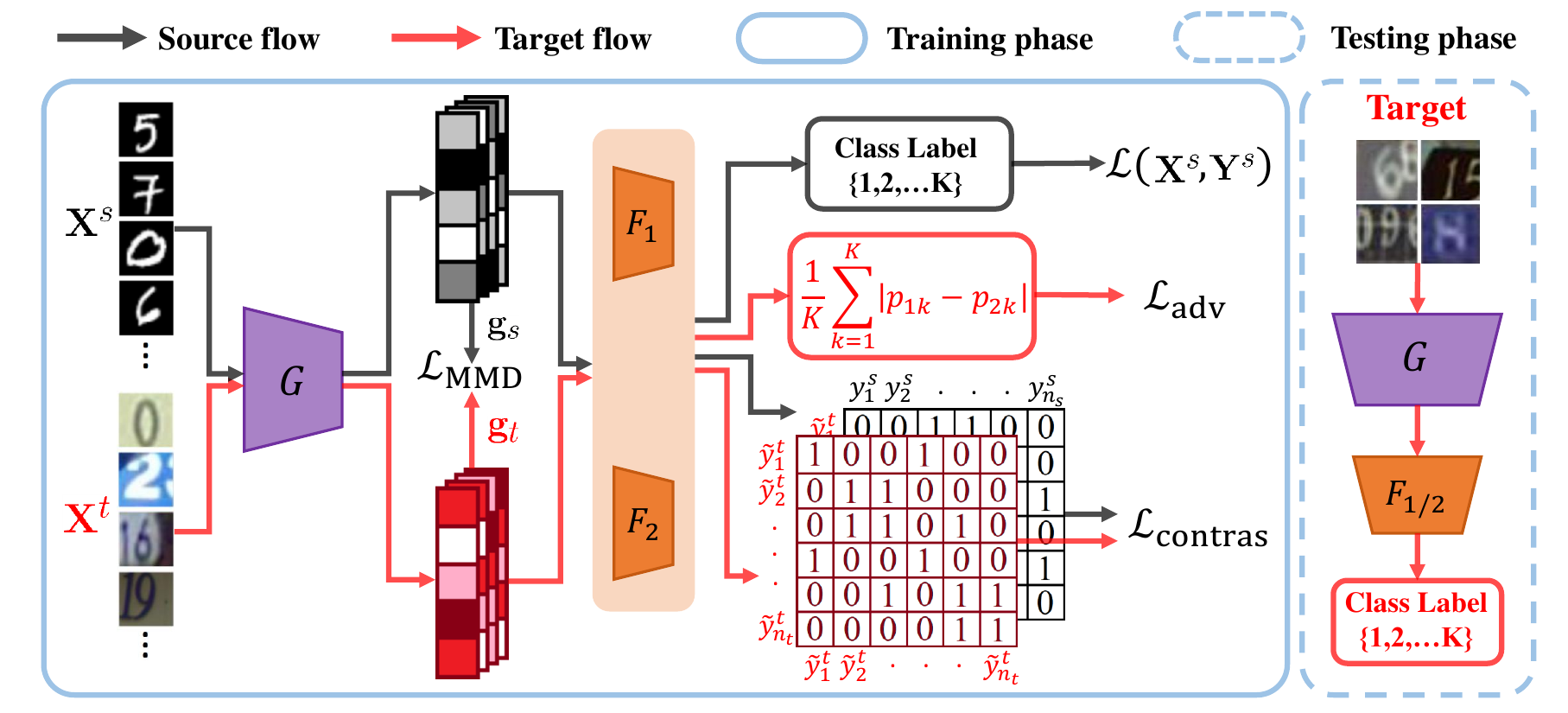}
    \caption{Framework of the proposed \model. The inputs are $\mathbf{X}^s$ with label $\mathbf{Y}^s$ from the source domain and unlabeled $\mathbf{X}^t$ from the target domain. The model contains a shared feature generator $G$ and two feature classifiers $F_1$ and $F_2$. $\mathcal{L}_{\text{MMD}}$ is calculated using the generated feature mean of the source and target, \textit{i.e.}, $g_s$ and $g_t$ respectively. $\mathcal{L}_{\text{adv}}$ is the classifier discrepancy calculated based on the probability outputs $p_1$ and $p_2$ of $F_1(G(\mathbf{X}^t))$ and $F_2(G(\mathbf{X}^t))$, respectively. $\mathcal{L}_{\text{contras}}$ is the contrastive loss calculated for both source-and-target and target-and-target samples.}
    \label{fig:model_framework}
\end{figure*}

\section{Approach}
Unsupervised domain adaptation seeks to generalize a learned model from a source domain to a target domain, the latter following a different (but related) data distribution from the former. Specifically, the source- and target-domain samples are denoted $\mathcal{S} =\{(\mathbf{x}_1^s, y_1^s),...,(\mathbf{x}_i^s, y_i^s),...,(\mathbf{x}^s_{N_s}, y^s_{N_s})\}$, and $\mathcal{T} =\{\mathbf{x}^t_1,...,\mathbf{x}^t_i,...,\mathbf{x}^t_{N_t}\}$, respectively, where $\mathbf{x}_i^s$ and $\mathbf{x}^t_i$ are the input, and $y_i^s \in \{1,2,...,K\}$ represents the data labels of $K$ classes in the source domain. The target domain shares the same label types as the source domain, but we possess no labeled examples from the target domain. We are interested in learning a deep network $G$ that reduces domain shift in the data distribution across $\mathcal{S}$ and $\mathcal{T}$, in order to make accurate predictions for $y_i^t$. We use the notation $(\mathbf{X}^{s},\mathbf{Y}^{s})$ to describe the source-domain samples and labels, and $\mathbf{X}^{t}$ for the unlabeled target-domain samples. 

Adversarial domain adaptation approaches such as \cite{saito2017maximum,mcdgau} achieve this goal via a two-step procedure: $i$) train a feature generator $G$ and the feature classifiers $F_1$, $F_2$ with the source-domain data, to ensure the generated features are class-conditional; $ii$) train $F_1$ and $F_2$ so that the prediction discrepancy between the two classifiers is maximized, and train $G$ to generate features that are distinctively separated. The maximum classifier discrepancy detects the target features that are far from the support of the source domain. As the generator tries to fool the classifiers ($i.e.$, minimizing the discrepancy), these target-domain features are enforced to be categorized and aligned with the source-domain features.

However, only measuring divergence between $F_1$ and $F_2$ can be considered first-order moment matching, which may be insufficient for adversarial training. Previous work also observed similar issues \cite{pmlr-v70-arora17a,Tsai_2018_CVPR}. We address this challenge by adding the Maximum Mean Discrepancy (MMD) loss, that matches the difference via higher-order moments. Also, the class alignment in existing UDA methods takes into account the intra-class domain discrepancy only, which makes it difficult to separate samples within the same class that are close to the decision boundary. Thus, in addition to the discrepancy loss, we also measure both intra- and inter-class discrepancy across domains. Specifically, we propose to minimize the distance among target-domain features that fall into the same class based on decision boundaries, and separate those features from different categories. During this process, ambiguous target features are simultaneously kept away from the decision boundaries and mapped into the high-density region, achieving better class alignment. 

\subsection{Global Alignment with MMD}
Following \cite{saito2017maximum}, we first train a feature generator $G(\cdot)$ and two classifiers $F_1(G(\cdot))$ and $F_2(G(\cdot))$ to minimize the softmax cross-entropy loss using the data from the labeled source domain $\mathcal{S}$, defined as:
\begin{align}
\label{eq:cross_entropy}
 \mathcal{L}(\mathbf{X}^{s},\mathbf{Y}^{s}) = & -\mathbb{E}_{(\mathbf{x}^{s},y^{s}) \sim (\mathbf{X}^{s},\mathbf{Y}^{s})} \Big[\sum_{k=1}^{K} \mathbbm{1}_{[k=y^{s}]} \log p_{1}(\mathbf{y}|\mathbf{x}^{s}) \nonumber \\
 &+\sum_{k=1}^{K}\mathbbm{1}_{[k=y^{s}]} \log p_{2}(\mathbf{y}|\mathbf{x}^{s})\Big]
\end{align}
where $p_1(\mathbf{y}|\mathbf{x})$ and $p_2(\mathbf{y}|\mathbf{x})$ are the probabilistic output of the two classifiers $F_1(G(\mathbf{x}))$ and $F_2(G(\mathbf{x}))$, respectively. 


In addition to \eqref{eq:cross_entropy}, we explicitly minimize the distance between the source and target feature distributions with MMD. The main idea of MMD is to estimate the distance between two distributions as the distance between sample means of the projected embeddings in a Hilbert space. Minimizing MMD is equivalent to minimizing all orders of moments \cite{gretton2012kernel}. In practice, the squared value of MMD is estimated with empirical kernel mean embeddings: 
\begin{equation}
\begin{split}
\mathcal{L}_{\text{MMD}}(\mathbf{X}^{s},\mathbf{X}^{t}) = \sum_{i=1}^{n_s}\sum_{j=1}^{n_t} k(\phi(\frac{\mathbf{g}_s}{||\mathbf{g}_s||}),\phi(\frac{\mathbf{g}_t}{||\mathbf{g}_t||})) \\
\mathbf{g}_s=\frac{1}{n_s}\sum_{i=1}^{n_s}G(\mathbf{x}_i^{s}),\quad \mathbf{g}_t=\frac{1}{n_t}\sum_{i=1}^{n_t}G(\mathbf{x}_i^{t})
\end{split}
\end{equation}
where $\phi(\cdot)$ is the kernel mapping, $\mathbf{g}_s\in\mathcal{R}^{n}$, $\mathbf{g}_t\in\mathcal{R}^{n}$, with $n_t$ and $n_s$ denoting the size of a training mini-batch of the data from the source domain $\mathcal{S}$ and the target domain $\mathcal{T}$, respectively; $||\cdot||$ denotes the $\ell_2$-norm.
With the MMD loss $\mathcal{L}_{\text{MMD}}$, the normalized features in the two domains are encouraged to be distributed identically, leading to better global domain alignment.

\subsection{Contrastively Smoothed Class Alignment}
\Paragraph{Discrepancy Loss}
The discrepancy loss represents the level of disagreement between the two feature classifiers in prediction for target-domain samples. Specifically, the discrepancy loss between $F_1$ and $F_2$ is defined as: 
\begin{equation}
       d(p_1(\mathbf{y}|\mathbf{x}),p_2(\mathbf{y}|\mathbf{x})) 
     = \frac{1}{K} \sum_{k=1}^K{\Big|p_{1_k}(\mathbf{y}| \mathbf{x})\!-\!p_{2_k}(\mathbf{y}|\mathbf{x})\Big|}
\end{equation}
where $|\cdot|$ denotes the $\ell_1$-norm, and $p_{1_k}(\cdot)$ and $p_{2_k}(\cdot)$ are the probability output of $p_1$ and $p_2$ for the $k$-th class, respectively. Accordingly, we can define the discrepancy loss over the target domain $\mathcal{T}$:
\begin{equation}
\mathcal{L}_{\text{adv}}(\mathbf{X}^t) = \mathbb{E}_{\mathbf{x}^{t} \sim \mathbf{X}^{t}}\big[d(p_1(\mathbf{y}|\mathbf{x}^t),p_2(\mathbf{y}|\mathbf{x}^t))\big]
\end{equation}
Adversarial training is conducted in the MCD~\cite{saito2017maximum} setup:
\begin{equation}
\begin{split}
& \mathop{\rm min}\limits_{F_1,F_2}  \mathcal{L}(\mathbf{X}^{s},\mathbf{Y}^{s}) - \lambda\mathcal{L}_{\text{adv}}(\mathbf{X}^t) \\
& \mathop{\rm min}\limits_{G} \mathcal{L}_{\text{adv}}(\mathbf{X}^t)
\end{split}
\end{equation}
where $\lambda$ is a hyper-parameter. Minimizing the discrepancy between the two classifiers $F_1$ and $F_2$ induces smoothness for the clearly classified target-domain features, while the region in the vacancy among the ambiguous ones remains non-smooth. Moreover, MCD only utilizes the unlabeled target-domain samples, while ignoring the labeled source-domain data when estimating the discrepancy.

\Paragraph{Contrastive Loss}
To further optimize $G$ to estimate the underlying label hypothesis of target-domain samples, we propose to measure the intra- and inter-class discrepancy across domains, conditional on class information. By using an indicator defined as $c(y,y^{\prime}) =
\begin{cases}
  1, & y = y^{\prime} \\
  0, & y \ne y^{\prime}
\end{cases}$, we define the contrastive loss between $\mathcal{S}$ and $\mathcal{T}$ as:
\begin{equation}
\label{eq:s2t}
\mathcal{L}^{\mathcal{S} \leftrightarrow \mathcal{T}}_{\text{contras}} = \sum_{\mathbf{x}^{s}_i \in \mathcal{S}, \mathbf{x}^{t}_j \in \mathcal{T}} L_{\text{dis}}(G(\mathbf{x}^{s}_i),G(\mathbf{x}^{t}_j),c(y^{s}_i,\widetilde{y}^{t}_j))
\end{equation}
where $L_{\text{dis}}$ is a distance measure (defined below), and $\widetilde{y}^{t}_j$ is the predicted target label for $\mathbf{x}^{t}_j$. Specifically, \eqref{eq:s2t} covers two types of class-aware domain discrepancies: $i$) intra-class
domain discrepancy ($y^{s}_i=\widetilde{y}^{t}_j$); and $ii$) inter-class domain discrepancy ($y^{s}_i \ne \widetilde{y}^{t}_j$). Note that $y^{s}_i$ is known, providing some supervision for parameter learning. 
Similarly, we can define the constrastive loss between $\mathcal{T}$ and $\mathcal{T}$ as:
\begin{equation}
\label{eq:t2t}
\mathcal{L}^{\mathcal{T} \leftrightarrow \mathcal{T}}_{\text{contras}} = \sum_{\mathbf{x}^{t}_i, \mathbf{x}^{t}_j \in \mathcal{T}} L_{\text{dis}}(G(\mathbf{x}^{t}_i),G(\mathbf{x}^{t}_j),c(\widetilde{y}^{t}_i,\widetilde{y}^{t}_j))
\end{equation}
To obtain the indicator $c(y,y^{\prime})$, estimated target label $\widetilde{y}^{t}_i$ is required. Specifically, for each data sample $\mathbf{x}_j^{t}$, a pseudo label is predicted based on the maximum posterior probability of the two classifiers: 
\begin{equation}\label{eq:pseudo}
\begin{split}
\widetilde{y}^{t}_j = \mathop{\arg\max}_{k\in \{ 1,2,...,K \} }\ & \Big\{p(F_1(G(\mathbf{x}_j^{t}))=k|\mathbf{\mathbf{x}})\\
& +p(F_2(G(\mathbf{x}_j^{t}))=k|\mathbf{x})\Big\}
\end{split}
\end{equation}
Ideally, based on the indicator, $L_{\text{dis}}$ should ensure the gathering of features that fall in the same class, while separating those in different categories. Following \cite{luo2017smooth}, we utilize contrastive
Siamese networks \cite{NIPS1993_769}, which can learn an invariant
mapping to a smooth and coherent feature space and perform
well in practice:
\begin{equation}\label{eq:L_dis}
L_{\text{dis}}=\left\{
\begin{aligned}
&||G(\mathbf{x}_i)-G(\mathbf{x}_j)||^2 &c_{ij}\!=\!1  \\
&\max(0, m\!-\!||G(\mathbf{x}_i)\!-\!G(\mathbf{x}_j)||)^2\!&c_{ij}\!=\!0
\end{aligned} \right.
\end{equation}
where $c_{ij}=c(y_i,y_j)$ and $m$ is a pre-defined margin. The margin loss constrains the neighboring features to be consistent. Based on the above definitions of source-and-target and target-and-target contrastive losses, the overall objective is:
\begin{equation}
\label{eq:overall}
\mathcal{L}_{\text{contras}}(\mathbf{X}^{s},\mathbf{Y}^{s},\mathbf{X}^{t}) = \mathcal{L}^{\mathcal{S} \leftrightarrow \mathcal{T}}_{\text{contras}} + \mathcal{L}^{\mathcal{T} \leftrightarrow \mathcal{T}}_{\text{contras}}
\end{equation}
Minimizing the contrastive loss $\mathcal{L}_{\text{contras}}$ encourages features in the same class to aggregate together while pushing unrelated pairs away from each other. In other words, the semantic feature approximation is enhanced to induce smoothness between data in the feature space. 

\begin{algorithm}[t]
\begin{algorithmic}[1]
\STATE \textbf{Input}: Source domain samples $\{\mathbf{x}^{s}_{i}, y^{s}_{i}\}$, and target domain samples $\{\mathbf{x}^{t}_{j}\}$. Hyper-parameters $\lambda_1$, $\lambda_2$, $\lambda_3$, and inner-loop iteration $\tau$ and $\delta$.
\STATE \textbf{Output}: Classifiers $F_1$ and $F_2$, and generator $G$.
\\\hrulefill
\FOR{$\text{$iter$ from 1 to $max\_iter$}$}
\STATE Sample a mini-batch of source samples $\left[\mathbf{x}_{i}^s, y_{i}^s\right]$ and target samples $\left[\mathbf{x}_{j}^t\right]$.
\vspace{2mm}
\STATE \textit{\textbf{\footnotesize \# Update both the generator and the classifiers}}
\STATE \text{Compute $\mathcal{L}(\mathbf{X}^{s},\mathbf{Y}^{s})$ on $\left[\mathbf{x}^{s}_{i}, y^{s}_{i}\right]$}.\\
\STATE \text{Compute $\mathcal{L}_{\text{MMD}}(\mathbf{X}^{s},\mathbf{X}^{t})$ on $\left[\mathbf{x}^{s}_{i}, \mathbf{x}^{t}_{j}\right]$}.\\
\STATE Update $G$, $F_1$ and $F_2$ using \eqref{eq:stepA}.\\
\vspace{2mm}
\STATE \textit{\textbf{\footnotesize \# Update the classifiers}}
\FOR{$\text{$inner\_loop\_iter_1$ from 1 to $\tau$}$}
\STATE \text{Compute $\mathcal{L}(\mathbf{X}^{s},\mathbf{Y}^{s})$ on $\left[\mathbf{x}^{s}_{i}, y^{s}_{i}\right]$}.\\
\STATE \text{Compute $\mathcal{L}_{\text{adv}}(\mathbf{X}^{t})$ on $\mathbf{x}^{t}_{j}$}.\\
\STATE Fix $G$, update $F_1$ and $F_2$ using \eqref{eq:stepB}.
\ENDFOR
\vspace{2mm}
\STATE \textit{\textbf{\footnotesize \# Update the feature generator}}
\FOR {$\text{$inner\_loop\_iter_2$ from 1 to $\delta$}$}
\STATE \text{Compute $\mathcal{L}_{\text{adv}}(\mathbf{X}^{t})$ on $\mathbf{x}^{t}_{j}$}.\\
\STATE \text{Compute $\mathcal{L}_{\text{contras}}(\mathbf{X}^{s},\mathbf{Y}^{s},\mathbf{X}^{t})$ on $\left[\mathbf{x}^{s}_{i}, y^{s}_{i},\mathbf{x}^{t}_{j}\right]$}.\\
\STATE Fix $F_1$ and $F_2$, update $G$ using \eqref{eq:stepC}.
\ENDFOR
\ENDFOR
\caption{Training procedure of \model.}
\label{alg:main}
\end{algorithmic}
\end{algorithm}

\subsection{Training Procedure}\label{subsec:train_detail}
We optimize $G$, $F_1$ and $F_2$ by combining all of the aforementioned losses, performed in an adversarial training manner. Specifically, we first train the classifiers $F_1$ and $F_2$ and the generator $G$ to minimize the objective:
\begin{equation}\label{eq:stepA}
\mathop{\rm min}\limits_{F_1,F_2,G}~ \mathcal{L}(\mathbf{X}^{s},\mathbf{Y}^{s}) + \lambda_1\mathcal{L}_{\text{MMD}}(\mathbf{X}^{s},\mathbf{X}^{t})
\end{equation}
We then train the classifiers $F_1$ and $F_2$ while keeping the generator $G$ fixed. The objective is:
\begin{equation}
\label{eq:stepB}
\mathop{\rm min}\limits_{F_1,F_2}~  \mathcal{L}(\mathbf{X}^{s},\mathbf{Y}^{s}) -\lambda_2\mathcal{L}_{\text{adv}}(\mathbf{X}^t) 
\end{equation}
Lastly, we train the generator $G$ with the following objective, while keeping both $F_1$ and $F_2$ fixed: 
\begin{equation}
\label{eq:stepC}
\begin{split}
\mathop{\rm min}\limits_{G}~  \lambda_2\mathcal{L}_{\text{adv}}(\mathbf{X}^t) 
+\lambda_3 \mathcal{L}_{\text{contras}}(\mathbf{X}^{s},\mathbf{Y}^{s},\mathbf{X}^{t})
\end{split}
\end{equation}
where $\lambda_1$, $\lambda_2$ and $\lambda_3$ are hyper-parameters that balance the different objectives.
These steps are repeated, with 
the full approach summarized in Algorithm \ref{alg:main}. In our experiments, the inner-loop iteration numbers $\tau$ and $\delta$ are both set to 2.

\Paragraph{Class-aware sampling}
When training with the contrastive loss, it is important to sample a mini-batch of data with all the classes, to allow \eqref{eq:overall} to be fully trained. Following \cite{kang2019contrastive}, we use a class-aware sampling strategy. Specifically, we randomly select a subset of each class, from which a mini-batch is sampled. Consequently, in each mini-batch, we are able to estimate the intra/inter-class discrepancy.

\Paragraph{Dynamic parameterization of $\lambda_3$}
In our implementation, we adapt a dynamic $\omega(t)$ to parameterize $\lambda_3$. We set $\omega(t) = \mathrm{exp}[-\theta(1-\frac{t}{\text{max-epochs}})]\lambda_3$, which is a Gaussian curve ranging from 0 to $\lambda_3$; this is employed to prevent unlabeled target features gathering in the early stage of training, as the pseudo labels might be unreliable.


\section{Experiments}
We evaluate the proposed model primarily on image datasets. To compare with MCD~\cite{saito2017maximum} as well as the state-of-the-art results in~\cite{shu2018dirt,kumar2018co}, we evaluate on the same datasets used in those studies: the digit datasets ($i.e.$, MNIST, MNISTM, Street View House Numbers (SVHN), and USPS), CIFAR-10, and STL-10. We also conduct experiments on the VisDA dataset, \textit{i.e.}, large-scale images. Our model can also be applied to non-visual domain adaptation tasks. Specifically, to show the flexibility of our model, we also evaluate it on the Amazon Reviews dataset.

For visual domain adaptation tasks, the proposed model is implemented based on VADA~\cite{shu2018dirt} and Co-DA~\cite{kumar2018co} to avoid any incidental difference caused by network architecture. However, different from these methods, our model does not require a discriminator, and only adopts the architecture for the feature generator $G$ and the classifier $F$.
Specifically, $G$ has 9 convolutional layers with several dropout, max-pool, Gaussian noise and global pool layers (details can be found in the Supplementary Material). Both $F_1$ and $F_2$ are one-layer MLPs. 
We also include instance normalization~\cite{shu2018dirt,ulyanov2016instance}, achieving superior results on several benchmarks.
For the VisDA dataset, we implemented our model based on Self-ensembling Domain Adaptation (SEDA)~\cite{french2017self}.
To compare with MCD and Contrastive Adaptation Network (CAN)~\cite{kang2019contrastive} (codebase not available) in both experiments, we re-implemented them using the exact architecture as our model.

\begin{figure}[t]
\centering
\subfigure[The Digit dataset.]{\label{fig:sample_digit} 
\includegraphics[width=0.47\textwidth]{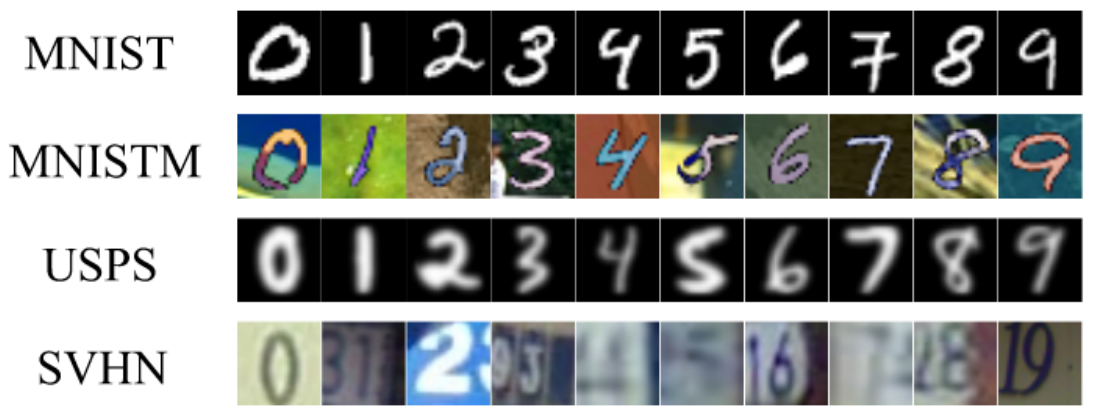}}
\subfigure[The CIFAR-10 dataset and the STL dataset.]{\label{fig:cifar_sample} 
\hspace{2mm}
\includegraphics[width=0.47\textwidth]{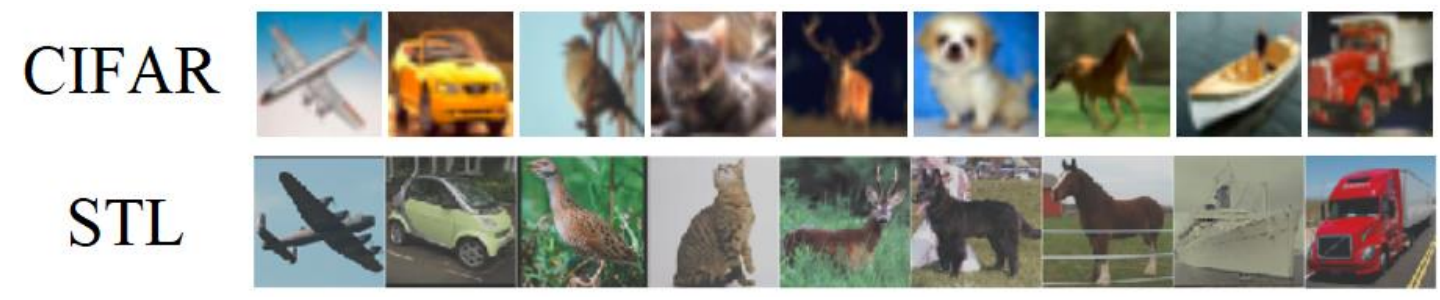}}
\caption{Sample images from the Digit, CIFAR-10 and STL datasets. Images from each column belong to the same class, while each row corresponds to a domain.}
\label{fig:samples_all}
\end{figure}

In addition to the aforementioned baseline models, we also include results from recently proposed unsupervised domain adaptation models. Note that standard domain adaptation methods (such as Transfer Component Analysis (TCA)~\cite{pan2011domain} and Subspace Alignment (SA)~\cite{fernando2013unsupervised}) are not included; these models only work on pre-extracted features, and are often not scalable to large datasets. Instead, we mainly compare our model with methods based on adversarial neural networks. 

For the non-visual task, we adopt a one-layer CNN structure from previous work \cite{D14-1181}. The feature generator $G$ consists of three components, including a 300-dimensional word embedding layer using GloVe \cite{pennington2014glove}, a one-layer CNN with ReLU, and a max-over-time pooling through which the final sentence representation is obtained. The classifiers $F_1$ and $F_2$ can be decomposed into one dropout layer and one fully connected output layer.

\subsection{Digit Datasets}
There are four types of digit images ($i.e.$, four domains). MNIST and USPS are both hand-written gray-scale images, with relatively small domain difference. MNISTM~\cite{ganin2016domain} is a dataset built upon MNIST by adding randomly colored image patches from BSD500 dataset~\cite{arbelaez2011contour}. SVHN includes colored images of street numbers. All images are rescaled to $32\times32\times3$. 
Sample images of all four digit datasets are presented in Figure~\ref{fig:sample_digit}.

\setlength{\tabcolsep}{8pt}
\begin{table*}[t]
  \centering
  \begin{tabular}{l|cccccc}
  \hline
     Source Domain & MNIST & SVHN & MNIST & MNIST & CIFAR & STL\\
     Target Domain & SVHN & MNIST & MNISTM & USPS & STL & CIFAR\\
    \hline \hline
      MMD~\cite{long2015learning} & - & 71.1 & 76.9 & 81.1 & - & - \\
      DANN~\cite{ganin2016domain} & 35.7 & 71.1 & 81.5 & 77.1 & - & - \\
      DSN~\cite{bousmalis2016domain} & 40.1 & 82.7 & 83.2 & 91.3 & - & - \\
      ATT~\cite{saito2017asymmetric} & 52.8 & 86.2 & 94.2 & - & - & - \\
    \hline
    \multicolumn{7}{c}{With Instance-Normalized Input:}\\
    \hline
      Souce-Only & 40.9 & 82.4 & 59.9 & 76.7 & 77.0 & 62.6 \\
      VADA~\cite{shu2018dirt} & 73.3 & 94.5 & 95.7 & - & 78.3 & 71.4 \\
      Co-DA~\cite{kumar2018co} & \textbf{81.3} & 98.6 & 97.3 & - & 80.3 & 74.5 \\
      MCD~\cite{saito2017maximum} & 68.7 & 96.2$^{\dagger}$ & 96.7 & 94.2$^{\dagger}$ & 78.1 & 69.2 \\
      SEDA \cite{french2017self} & 37.5 & \textbf{99.2} & - & 98.2 & 80.1 & 74.2 \\
      CAN~\cite{kang2019contrastive} & 67.1 & 94.8 & 96.2 & 97.5 & 77.3 & 70.4 \\
      \model & 80.7 & 98.7 & \textbf{98.9} & \textbf{99.3} & \textbf{81.7} & \textbf{75.2} \\
  \hline
  \end{tabular}
  \vspace{2.5mm}
  \caption{Results on visual domain adaptation tasks. Source-Only corresponds to training a classifier in the source domain and applying it directly to the target domain, without any adaptation. Models with instance-normalized input are implemented using the same network architecture. Results with $\dagger$ are reported in~\cite{saito2017maximum}.} 
  \label{tab:image_result}
\end{table*}

\Paragraph{MNIST$\rightarrow$SVHN}
While both MNIST and SVHN include images of digits, there exists a large domain gap between these two datasets. As gray-scale handwritten digits, MNIST has much lower dimensionality than SVHN, which contains cropped street-view images of house numbers. Specifically, each image from SVHN has a colored background, which sometimes contains multiple digits, and might be blurry. This makes MNIST$\rightarrow$SVHN a much harder adaptation task than other digit datasets. It is shown recently in~\cite{shu2018dirt} that instance normalization allows the classifier to be invariant to channel-wide scaling and shifting of the input pixel intensities, greatly improving the adaptation performance on MNIST$\rightarrow$SVHN (73.3$\%$). With instance normalization, our proposed \model~achieves test accuracy of 80.7$\%$, as shown in Table~\ref{tab:image_result}, competitive with state-of-the-art results from~\cite{kumar2018co}. 

\begin{figure*}[t]
\centering
\hspace{-3mm}
\subfigure[MCD M$\rightarrow$S]{\label{fig:tsne_svhn_mcd} 
\includegraphics[width=0.25\textwidth]{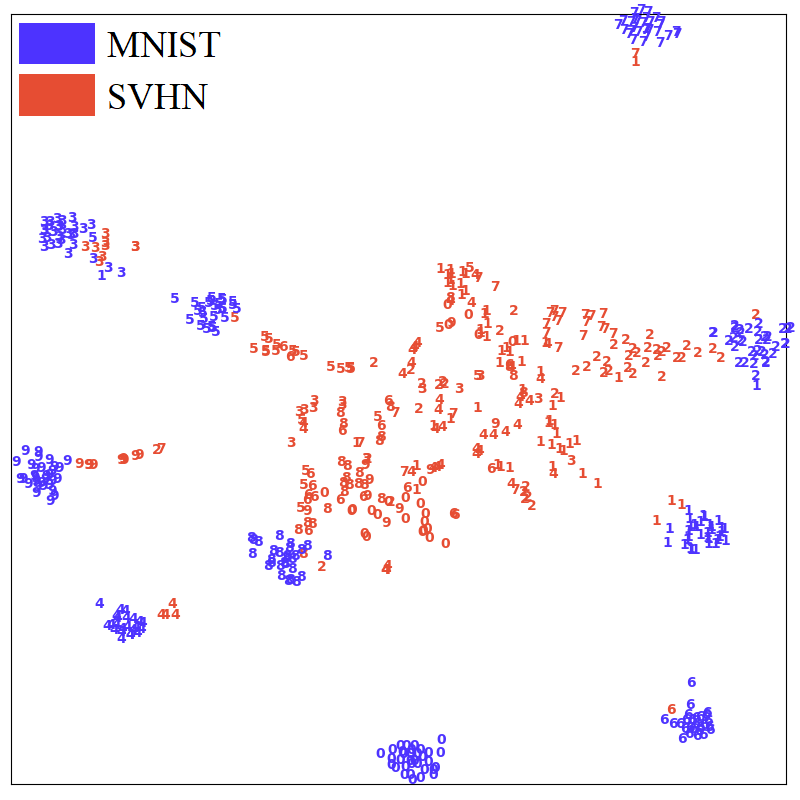}}
\hspace{-3mm}
\subfigure[\model~M$\rightarrow$S]{\label{fig:tsne_svhn_ours} 
\includegraphics[width=0.25\textwidth]{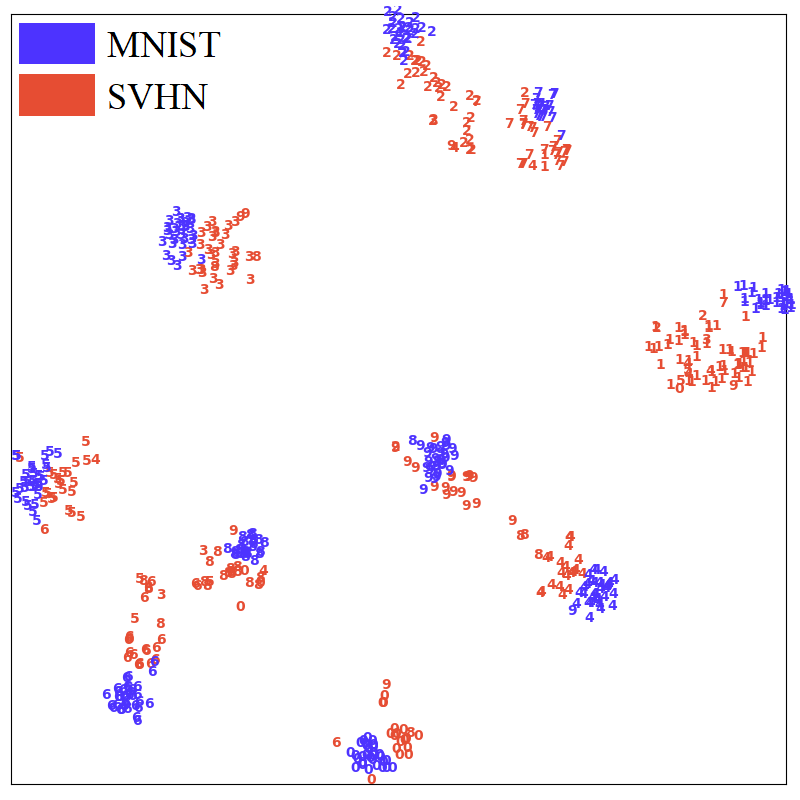}}
\hspace{-3mm}
\subfigure[MCD S$\rightarrow$M]{\label{fig:tsne_mnist_mcd} 
\includegraphics[width=0.25\textwidth]{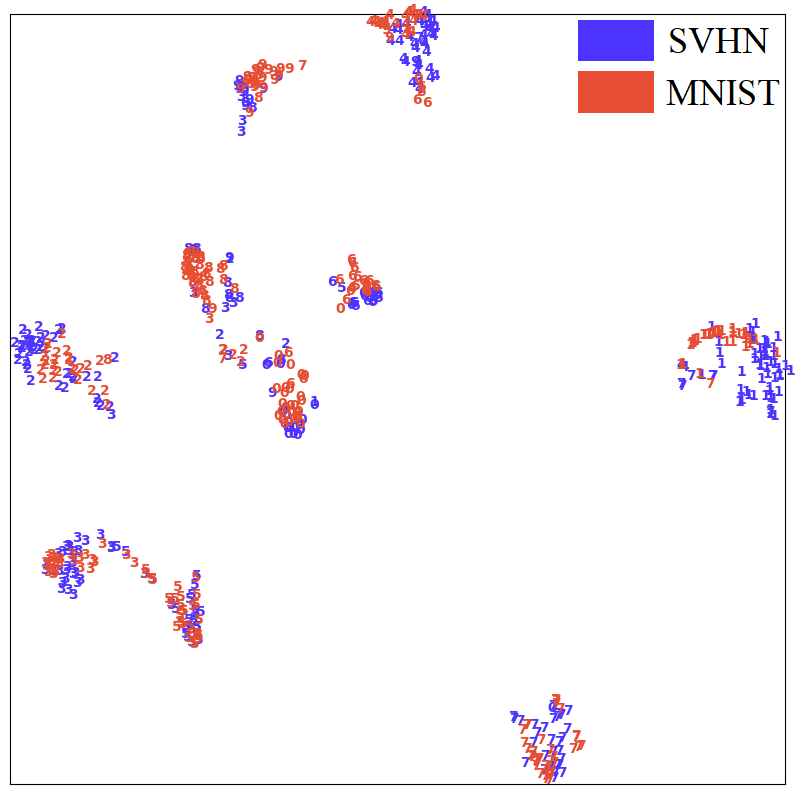}}
\hspace{-3mm}
\subfigure[CoSCA S$\rightarrow$M]{\label{fig:tsne_mnist_ours} 
\includegraphics[width=0.25\textwidth]{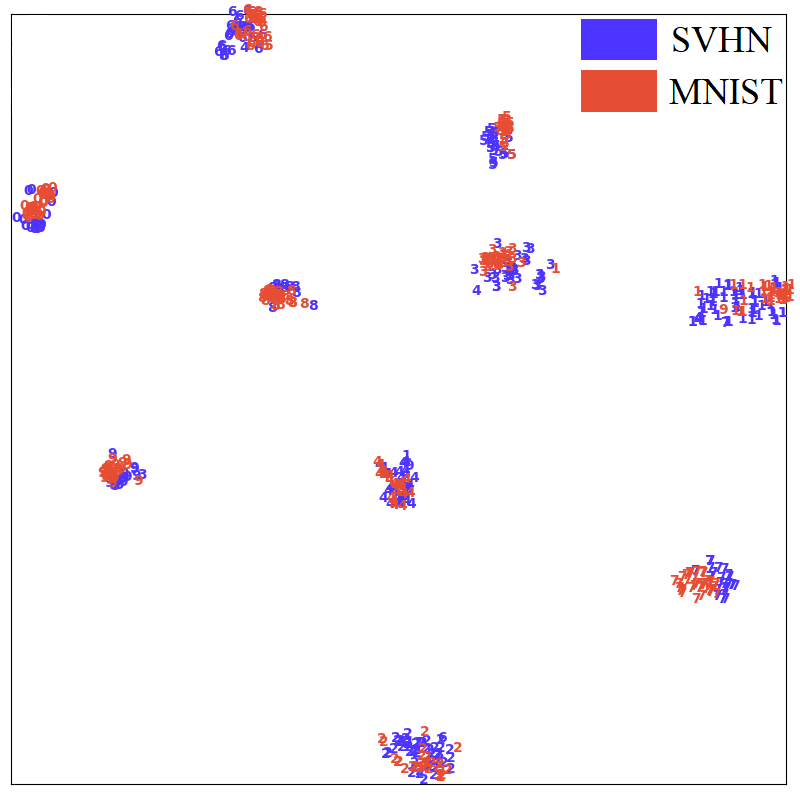}}
\vspace{3mm}

\hspace{-3mm}
\subfigure[MCD C$\rightarrow$STL]{\label{fig:tsne_stl_mcd} 
\includegraphics[width=0.25\textwidth]{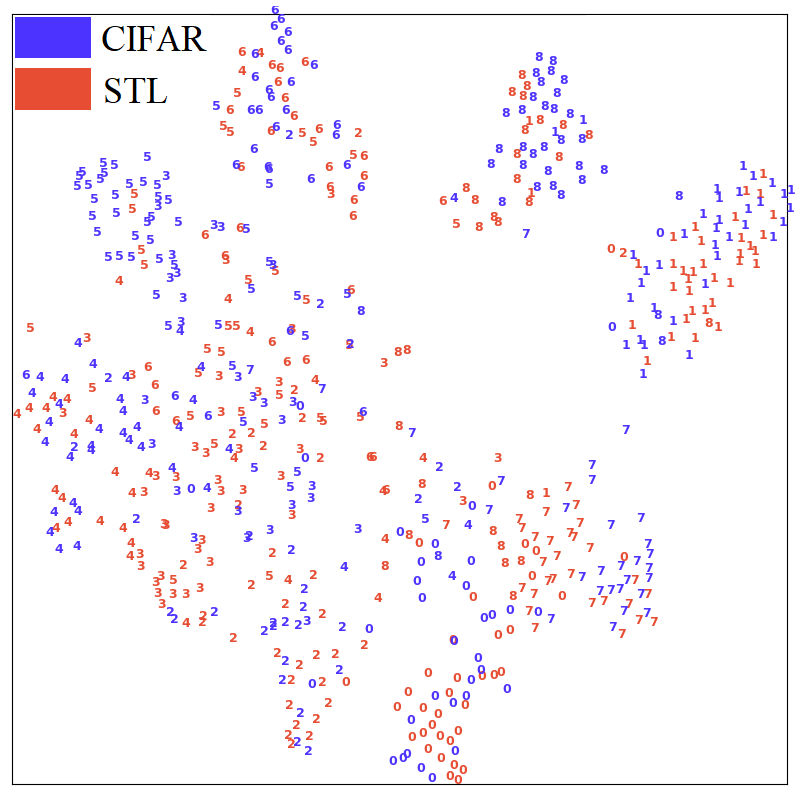}}
\hspace{-3mm}
\subfigure[CoSCA C$\rightarrow$STL]{\label{fig:tsne_stl_ours}
\includegraphics[width=0.25\textwidth]{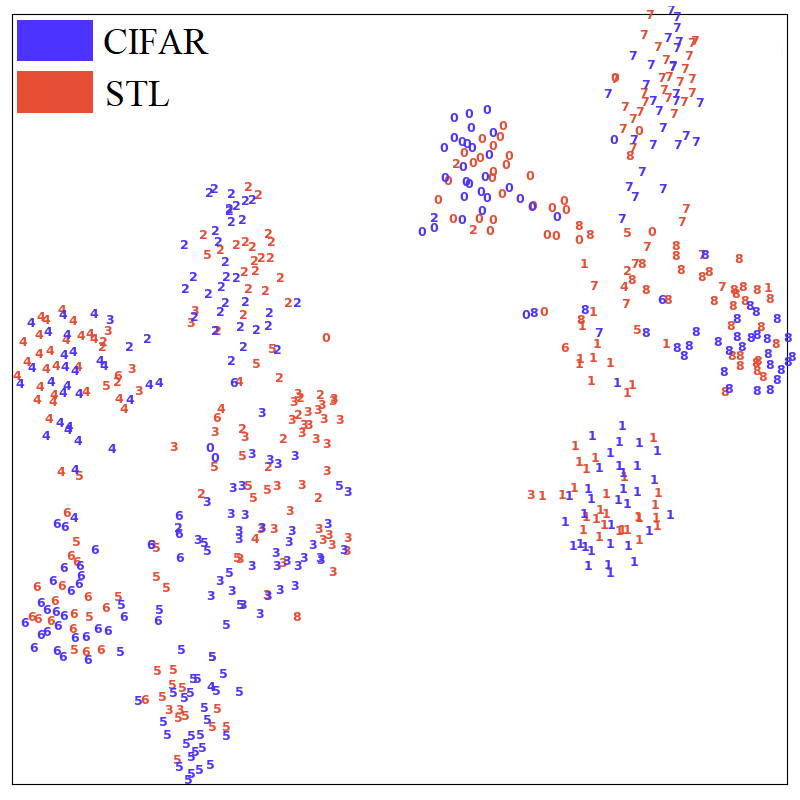}}
\hspace{-3mm}
\subfigure[MCD STL$\rightarrow$C]{\label{fig:tsne_cifar_mcd} 
\includegraphics[width=0.25\textwidth]{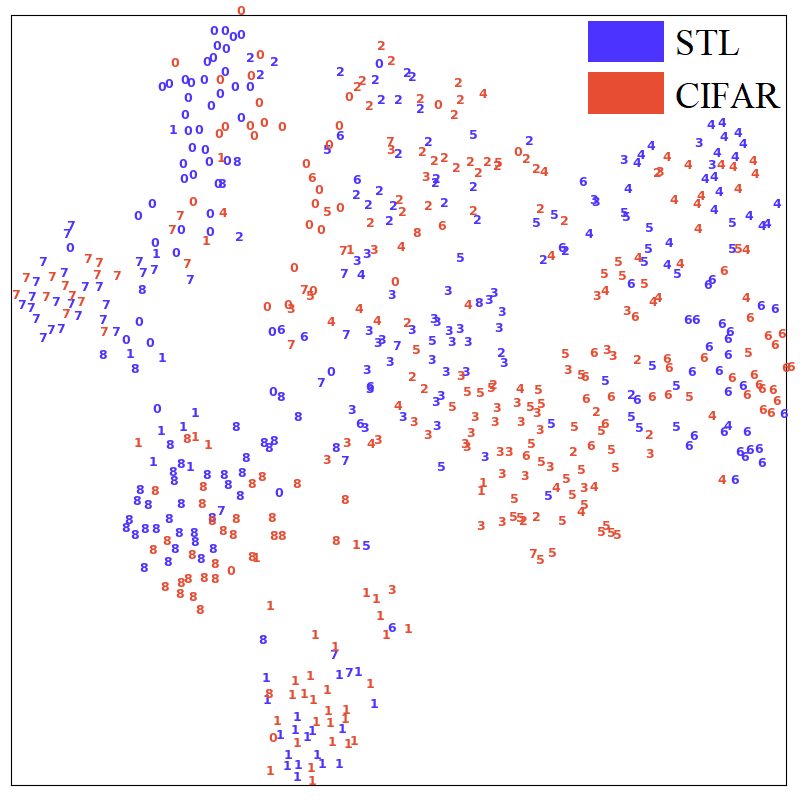}}
\hspace{-3mm}
\subfigure[\model~STL$\rightarrow$C]{\label{fig:tsne_cifar_ours}
\includegraphics[width=0.25\textwidth]{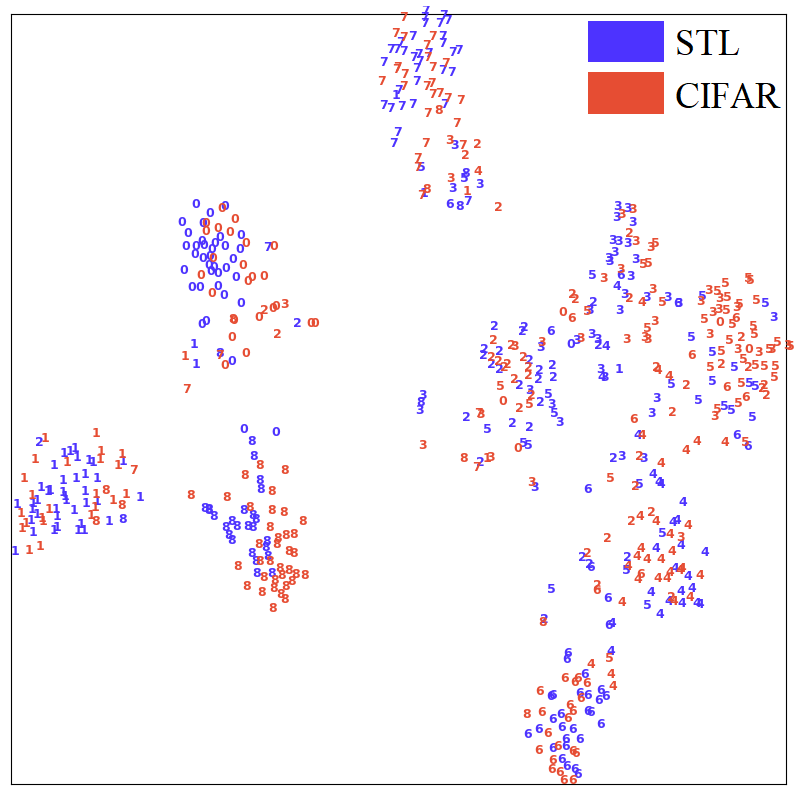}}
\caption{t-SNE embedding of the features $G(x)$ for MNIST (M) $\rightarrow$ SVHN (S) and STL $\rightarrow$ CIFAR (C). Color indicates domain, and the digit number is the label. The ideal situation is to mix the two colors with the same label, representing domain-invariant features. The t-SNE plots for the other datasets are provided in the Supplementary Material.}
\label{fig:tsne_all}
\end{figure*}

Notice that MCD does not provide adequate performance. Figure~\ref{fig:tsne_svhn_mcd} plots the t-SNE embedding of the features learned by MCD. Domains are indicated by different colors, and classes are indicated by different digit numbers. MCD fails to align the features of the two domains globally due to the large domain gap. In other words, the maximized discrepancy provides too many ambiguous target-domain samples. As a result, the feature generator may not be able to properly align them with the source-domain samples. In comparison, as shown in Figure~\ref{fig:tsne_svhn_ours}, \model~utilizes the MMD between the source and the target domain features, thus maintaining a better global domain alignment. With further smoothed class-conditional adaptation, it outperforms MCD. 

\Paragraph{SVHN$\rightarrow$MNIST}
Classification with the MNIST dataset is easier than others. As shown in Table~\ref{tab:image_result}, source-only achieves  82.4$\%$ on SVHN$\rightarrow$MNIST with instance normalization. Therefore, even with the same amount of domain difference, performance on SVHN$\rightarrow$MNIST is much better than MNIST$\rightarrow$SVHN across all compared models. The test accuracy of our model achieves 98.7$\%$.

\Paragraph{MNIST$\rightarrow$MNISTM}
Since MNISTM is a colored version of MNIST, there exists a one-to-one matching between the two datasets, $i.e.$, a domain adaptation model would perform well as long as domain-invariant features are properly extracted. \model~provides better results than Co-DA, yielding a test accuracy of 98.9$\%$.

\Paragraph{MNIST$\rightarrow$USPS}
Evaluation on MNIST and USPS datasets is also conducted to compare our model with other baselines. The proposed method achieves an excellent result of 99.3$\%$.




\subsection{CIFAR-10 and STL-10 Datasets}
CIFAR-10 and STL-10 are both 10-class datasets, with each image containing an animal or a type of transportation. Images from each class are much more diverse than the digit datasets, with higher intrinsic dimensionality, which makes it a harder domain adaptation task. There are 9 overlapping classes between these two datasets.
Figure~\ref{fig:cifar_sample} shows some sample images from each class.
CIFAR provides images of size 32$\times$32 and a large training set of 50,000 image samples, while STL contains higher quality images of size 96$\times$96, but with a much smaller training set of 5,000 samples. Following~\cite{french2017self,shu2018dirt,kumar2018co}, we remove non-overlapping classes from these two datasets and resize the images from STL to 32$\times$32. 

STL$\rightarrow$CIFAR is more difficult than CIFAR$\rightarrow$STL, due to the small training set in STL. For the latter, the source-only model with no adaptation involved achieves an accuracy of 77.0$\%$. With adaptation, the margin-of-improvement is relatively small, while \model~provides the best improvement of 4.7$\%$ among all the models (Table \ref{tab:image_result}). For STL$\rightarrow$CIFAR, our model yields a 12.6$\%$ margin-of-improvement and an accuracy of 75.2$\%$. Figures~\ref{fig:tsne_stl_mcd}, \ref{fig:tsne_stl_ours}, \ref{fig:tsne_cifar_mcd}, and~\ref{fig:tsne_cifar_ours} provide t-SNE plots for MCD and our model, respectively, which shows our model achieves much better alignment for each class.

\setlength{\tabcolsep}{2pt}
\begin{table*}[t]
  \centering
  \begin{tabular}{l|cccccccccccc|c}
  \toprule
    Model & plane & bcycl & bus & car & horse & knife & mcycl & person & plant & sktbrd & train & truck & mean\\
    \midrule \midrule
    Source Only & 55.1 & 53.3 & 61.9 & 59.1 & 80.6 & 17.9 & 79.7 & 31.2 & 81.0 & 26.5 & 73.5 & 8.5 & 52.4$^{\dagger}$ \\
    MMD~\cite{long2015learning} & 87.1 & 63.0 & 76.5 & 42.0 & 90.3 & 42.9 & 85.9 & 53.1 & 49.7 & 36.3 & 85.8 & 20.7 & 61.1$^{\dagger}$ \\
    DANN~\cite{ganin2016domain} & 81.9 & 77.7 & 82.8 & 44.3 &  81.2 & 29.5 & 65.1 & 28.6 & 51.9 & 54.6 & 82.8 & 7.8 & 57.4$^{\dagger}$ \\
    \midrule
    MCD~\cite{saito2017maximum} & 89.1 & 80.8 & 82.9 & 70.9 & 91.6 & 56.5 & 89.5 & 79.3 & 90.9 & 76.1 & 88.3 & 29.3 & 77.1 \\
    CAN~\cite{kang2019contrastive} & 91.4 & 78.9 & 79.1 & 72.8 & 93.2 & 63.4 & 82.4 & 68.6 & 93.2 & 88.3 & 84.1 & \textbf{39.2} & 77.9\\
    SEDA~\cite{french2017self} & 95.3 & 87.1 & 84.2 & 58.3 & 94.4 & \textbf{89.6} & 87.9 & 79.1 & 92.8 & \textbf{91.3} & \textbf{89.6} & 37.4 & 82.2 \\
    \model & \textbf{95.7} & \textbf{87.4} & \textbf{85.7} & \textbf{73.5} & \textbf{95.3} & 72.8 & \textbf{91.5} & \textbf{84.8} & \textbf{94.6} & 87.9 & 87.9 & 36.8 & \textbf{82.9} \\
  \bottomrule
  \end{tabular}
  \vspace{2.5mm}
  \caption{Test accuracy of ResNet101 model fine-tuned on the VisDA dataset. Results with $\dagger$ are reported in~\cite{saito2017maximum}, while the others are implemented using the same network architecture.}
  \label{tab:visda_result}
\end{table*}

\subsection{VisDA Dataset}
The VisDA dataset is a large-scale image dataset that evaluates the adaptation from synthetic-object to real-object images. Images from the source domain are synthetic renderings of 3D models from different angles and lighting conditions. There are 152,397 image samples in the source domain, and 55,388 image samples in the target domain. 
The image size, after rescaling as in \cite{saito2017maximum}, is $224\times224\times3$. 
A model architecture with ResNet101~\cite{he2016deep} pre-trained on Imagenet is required. 
There are 12 different object categories in VisDA, shared by the source and the target domains.

Table~\ref{tab:visda_result} shows the test accuracy of different models in all object classes.
The class-aware methods, namely MCD~\cite{saito2017maximum}, SEDA~\cite{french2017self} and our proposed \model, outperform the source only model in all categories. In comparison, the methods that are mainly based on distribution matching do not perform well in some of the categories.
\model~outperforms MCD, showing the effectiveness of contrastive loss and MMD global alignment. In addition, it performs better than SEDA in most categories, demonstrating its robustness in handling large scale images.

\setlength{\tabcolsep}{5pt}
\begin{table*}[t]
  \centering
  \begin{tabular}{c|c|c|c|c|c|c}
  \hline 
    & Source-Only &  DANN \cite{ganin2016domain} & PBLM \cite{N18-1112}  & MCD \cite{saito2017maximum} & DAS \cite{D18-1383} & \model \\
   \hline \hline
     Accuracy & 79.13 & 80.29$^{\dagger}$ & 80.40$^{\dagger}$ & 81.35 & 81.96$^{\dagger}$ & \textbf{83.17} \\
  \hline
  \end{tabular}
  \vspace{2.5mm}
  \caption{Results on the Amazon Reviews dataset. Results with $\dagger$ are reported by \cite{D18-1383,N18-1112}.} \label{tab:text_result}
\end{table*}

\subsection{Amazon Reviews Dataset}
We also evaluate \model~on the Amazon Reviews dataset collected by Blitzer \textit{et al.} \cite{blitzer7domain}. It contains reviews from several different domains, with 1000 positive and 1000 negative reviews in each domain. 

Table~\ref{tab:text_result} shows the average classification accuracy of different methods, including DANN \cite{ganin2016domain} DAS \cite{D18-1383} and PBLM \cite{N18-1112}. We use the same model architecture and parameter setting for MCD and the source-only model. Results show that the proposed \model~outperforms all other methods. Specifically, it improves the performance from test accuracy of 81.96$\%$ to 83.17$\%$, when compared to the state-of-the-art method DAS. MCD achieves 81.35$\%$, which is also outperformed by \model.


\subsection{Ablation Study}
To further demonstrate the improvement of \model~over MCD~\cite{saito2017maximum}, we conduct an ablation study. Specifically, with the same network architecture and setup, we compare model performance among 1) MCD, 2) MCD with smooth alignment (MCD+Contras), 3) MCD with global alighnment (MCD+MMD), and 4) \model, to validate the effectiveness of adding contrastive loss $\mathcal{L}_{\text{contras}}$ and MMD loss $\mathcal{L}_{\text{MMD}}$ to MCD. As MCD has already achieved superior performance on some of the benchmark datasets, we mainly choose those tasks on which MCD does not perform very well, in order to better analyze the margin of improvement. Therefore, MNIST$\rightarrow$SVHN, STL$\rightarrow$CIFAR and Amazon Reviews are selected for this experiment (Table~\ref{tab:ablation_result}).

\setlength{\tabcolsep}{12pt}
\begin{table}[t]
  \centering
  \begin{tabular}{l|ccc}
  \hline 
      & MNIST & STL & Amazon \\
     Model & SVHN & CIFAR & Reviews \\
  \hline \hline
     MCD~\cite{saito2017maximum} & 68.7 & 69.2 & 81.35 \\
     MCD+MMD  & 72.1 & 70.2 & 81.73 \\
     MCD+Contras & 75.9 & 73.4 & 82.56 \\
     \model & \textbf{80.7} & \textbf{75.2} & \textbf{83.17} \\
  \hline
  \end{tabular}
  \vspace{2.5mm}
  \caption{Ablation study on \model~with different variations of MCD on MNIST$\rightarrow$SVHN, STL$\rightarrow$CIFAR, and Amazon Reviews.}
  \label{tab:ablation_result}
\end{table}

\begin{figure*}[t]
\centering
\hspace{-0.025\textwidth}
\subfigure[Validation accuracy vs number of iterations for MNIST$\rightarrow$SVHN.]{\label{fig:ablation_svhn}
\includegraphics[width=0.47\textwidth]{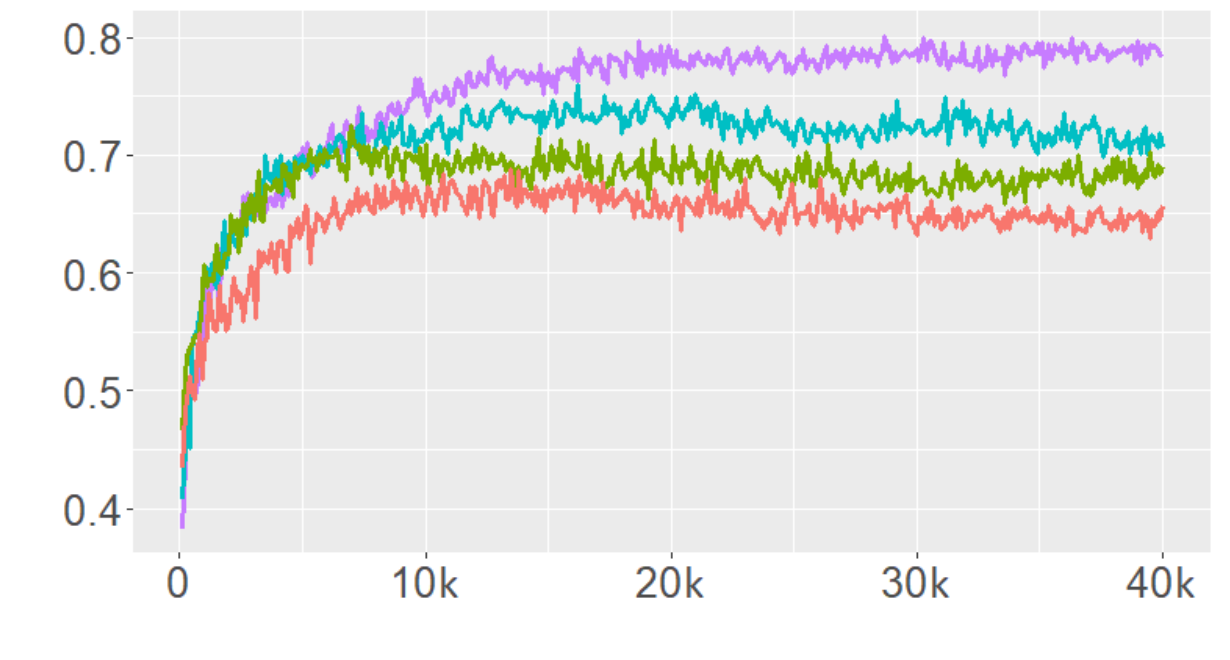}}
\hspace{0.05\textwidth}
\subfigure[Validation accuracy vs number of iterations for STL$\rightarrow$CIFAR.]{\label{fig:ablation_cifar}
\includegraphics[width=0.47\textwidth]{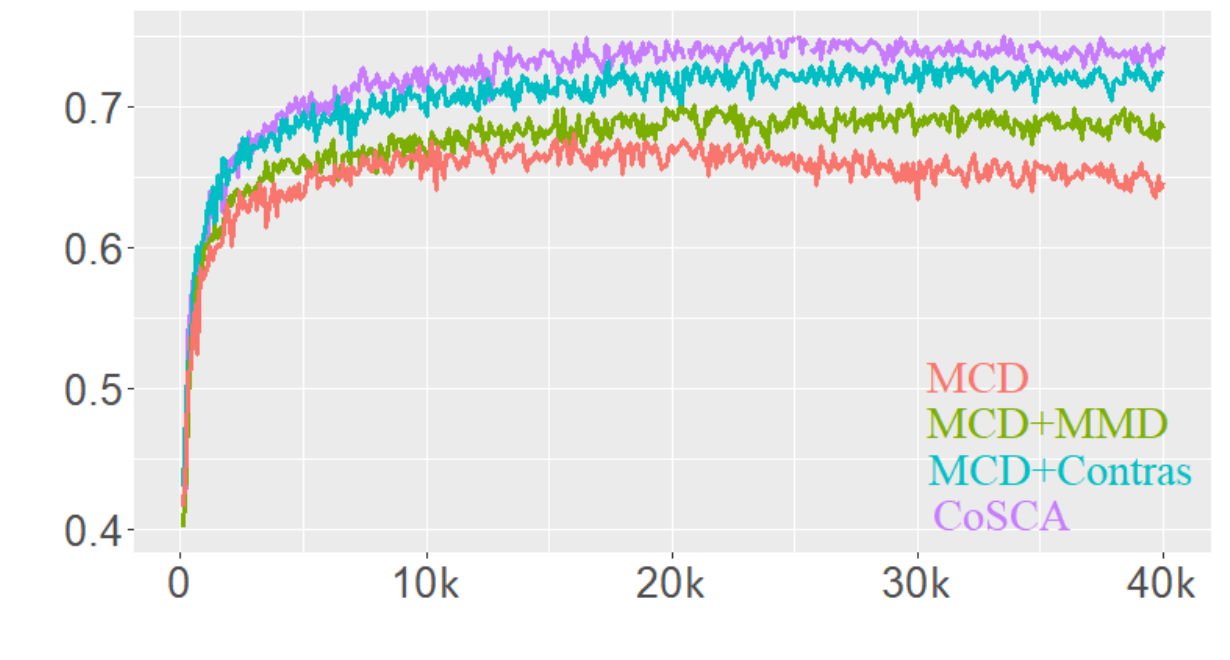}}
\vspace{-2mm}
\caption{Ablation study on \model~and different variations of MCD.}
\label{fig:ablation_all}
\end{figure*}

\Paragraph{Effect of Contrastive Alignment}
We compare \model~with MCD as well as its few variations, to validate the effectiveness of the proposed contrastive alignment. Table \ref{tab:ablation_result} provides the test accuracy for every model across the selected benchmark datasets. For MNIST$\rightarrow$SVHN, MCD+Contrastive outperforms MCD by 7.2$\%$. For STL$\rightarrow$CIFAR and Amazon Reviews, the margin of improvement is 4.2$\%$ and 1.21$\%$, respectively (less significant than MNIST$\rightarrow$SVHN, possibly due to the smaller domain difference). Note that the results of MCD+Contras are still worse than \model, demonstrating the effectiveness of the global domain alignment and the framework design of our model.

\Paragraph{Effect of MMD}
We further investigate how the MMD loss can impact the performance of our proposed \model. Specifically, for MNIST$\rightarrow$SVHN, MCD+MMD achieves a test accuracy of 72.1$\%$, only lifting the original result of MCD by 3.4$\%$. For STL$\rightarrow$CIFAR and Amazon Reviews, the margin-of-improvement is 1.0$\%$ and 0.38$\%$, respectively. While this validates the effectiveness of having global alignment in the MCD framework, the improvement is small. Without a smoothed class-conditional alignment, MCD still encounters misclassified target features during training, leading to a sub-optimal solution. Notice that when comparing \model~with MCD+Contras, the improvement for MNIST$\rightarrow$SVHN is significant (as shown in Figure~\ref{fig:ablation_svhn}), with validation accuracy and training stability enhanced. This demonstrates the importance of global alignment when there exists a large domain difference.


\section{Conclusions}
We have proposed Contrastively Smoothed Class Alignment (\model) for the UDA problem, by explicitly combining intra-class and inter-class domain discrepancy and optimizing class alignment through end-to-end mini-batch training. 
Experiments on several benchmarks demonstrate that our model can outperform state-of-the-art baselines. 
Our experimental analysis shows that \model~learns more discriminative target domain features, and the introduced MMD feature matching improves the global domain alignment. 
For future work, we will extend our model to other domain-adaptation tasks. Another direction to explore concerns development of a theoretical interpretation of contrastive learning for domain adaptation, particularly characterizing its effects on the alignment of source and target domain feature distributions.

\section*{Acknowledgements} 
The authors would like to thank the anonymous reviewers for their insightful comments. The research at Duke University was supported in part by DARPA, DOE, NIH, NSF and ONR.

\bibliographystyle{splncs}
\bibliography{egbib}

\end{document}


\pagestyle{headings}
\mainmatter

\def\ACCV20SubNumber{721}  

\title{Supplementary Material: Contrastively Smoothed Class Alignment for Unsupervised Domain Adaptation} 
\titlerunning{Supplementary Material}
%

\author{}
\institute{}
\authorrunning{S. Dai et al.}
%

\maketitle

\vspace{13.5mm}

\section{Model Architecture}
\noindent The  model architecture for the image datasets is listed in the following.
\begin{table}[h!]
\hspace{-7.5mm}
	\small
	\begin{tabular}{c|c}
		\toprule
		Feature Classifier & Feature Generator\\
		\midrule
		Input feature $G(X)$ & Input $X$  \\
		\midrule
		& $3\times3$ conv. 64 (96) lReLU, stride 1\\
	    & $3\times3$ conv. 64 (96) lReLU, stride 1\\
		& $3\times3$ conv. 64 (96) lReLU, stride 1\\
		& $2\times2$ max pool, stride 2, dropout, $p = 0.5$, Gaussian noise, $\sigma=1$\\
		
		& $3\times3$ conv. 64 (192) lReLU, stride 1\\
		& $3\times3$ conv. 64 (192) lReLU, stride 1\\
		MLP output $F(G(X))$ with shape 10 & $3\times3$ conv. 64 (192) lReLU, stride 1\\                
		& $2\times2$ max pool, stride 2, dropout, $p = 0.5$, Gaussian noise, $\sigma=1$\\
		
		& $3\times3$ conv. 64 (192) lReLU, stride 1\\
		& $3\times3$ conv. 64 (192) lReLU, stride 1\\
		& $3\times3$ conv. 64 (192) lReLU, stride 1\\
		& global average pool, output feature $G(X)$ with shape 64 (192)\\
		\bottomrule
	\end{tabular} 
	\vspace{3mm}
	\caption{Model architecture for the visual domain adaptation experiments. Numbers in the $(\cdot)$ are for CIFAR$\rightarrow$STL and STL$\rightarrow$CIFAR.}
	\label{Table:architecture}
	\vspace{-5mm}
\end{table}

\section{Hyper-parameter Setup}
\noindent The hyper-parameter setups for both the visual and non-visual datasets are listed in the following.
\setlength{\tabcolsep}{12pt}
\begin{table}[!ht]
	\vskip 0.05in
	\centering
	\small
	\begin{tabular}{l|ccc}
		\toprule
		Task & $\lambda_1$ for $\mathcal{L}_{\text{MMD}}$ & $\lambda_2$ for $\mathcal{L}_{\text{adv}}$ & $\lambda_3$ for $\mathcal{L}_{\text{contras}}$\\
		\midrule
		Digits & 10.0 & 0.1 & 0.2 \\
		CIFAR$\rightarrow$STL & 5.0 & 0.1 & 0.4 \\
		STL$\rightarrow$CIFAR & 5.0 & 0.1 & 0.2 \\
		Amazon Reviews & 4.0 & 0.1 & 0.8 \\
		\bottomrule
	\end{tabular} 
	\vspace{3mm}
	\caption{Hyper-parameter setup for visual and non-visual domain adaptation experiments.}
	\label{Table:hyper_parameter}
\end{table}

\section{Additional Experimental Results}
For fair comparison, the results on VisDA dataset in the main paper is reported based on ResNet101. Some of the results are reported based on ResNet152 originally, and therefore, we include them in Table \ref{tab:visda_result_2} as follows.

\begin{figure}[!t]
\centering
\hspace{-3mm}
\subfigure[MCD M$\rightarrow$U]{\label{fig:tsne_usps_mcd} 
\includegraphics[width=0.25\textwidth]{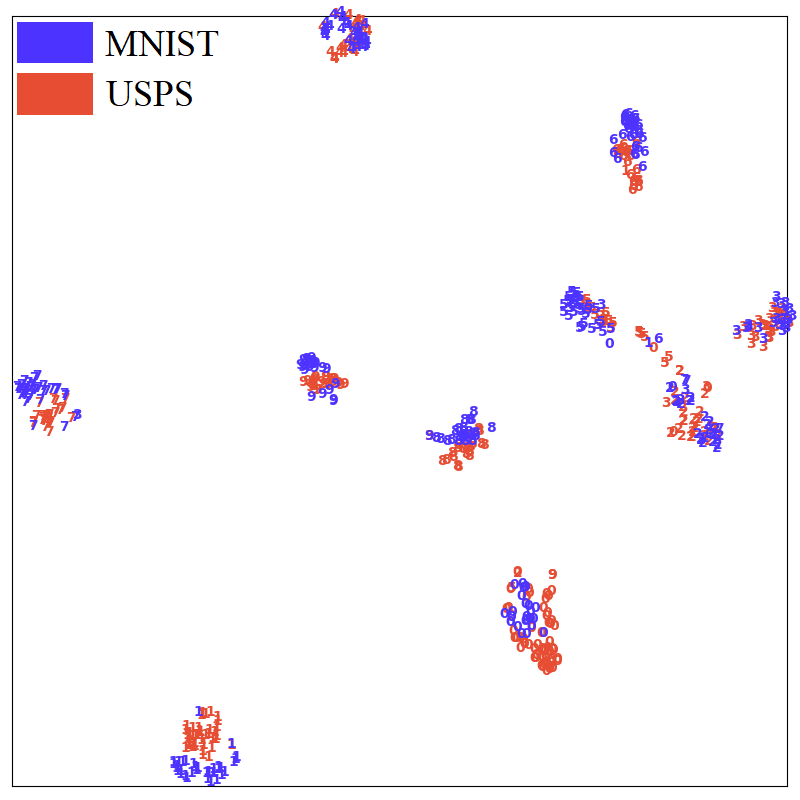}}
\hspace{-3mm}
\subfigure[CoSCA M$\rightarrow$U]{\label{fig:tsne_usps_ours} 
\includegraphics[width=0.25\textwidth]{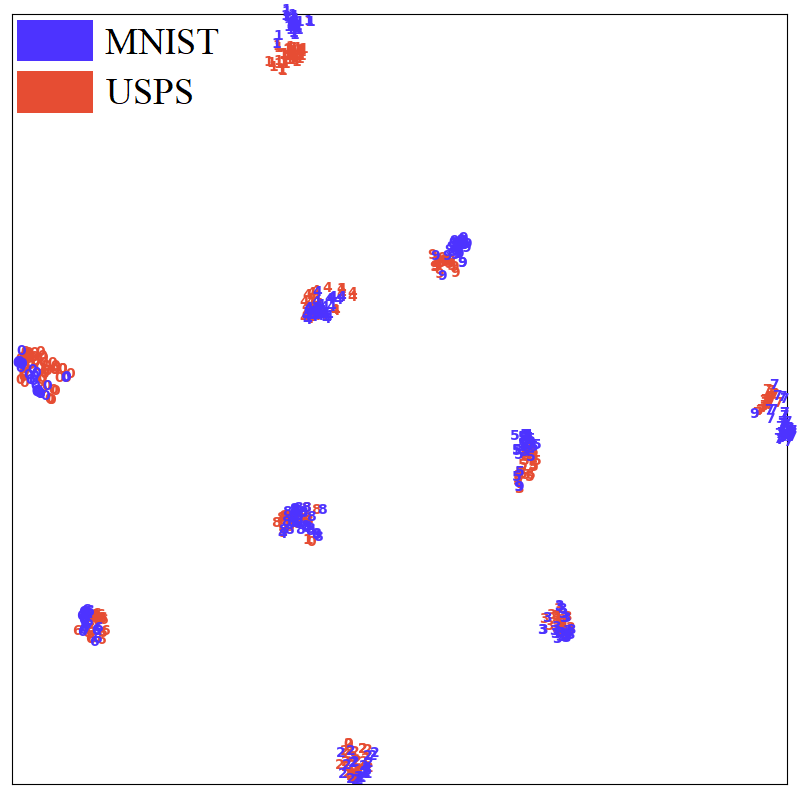}}
\hspace{-3mm}
\subfigure[MCD M$\rightarrow$MM]{\label{fig:tsne_mnistm_mcd} 
\includegraphics[width=0.25\textwidth]{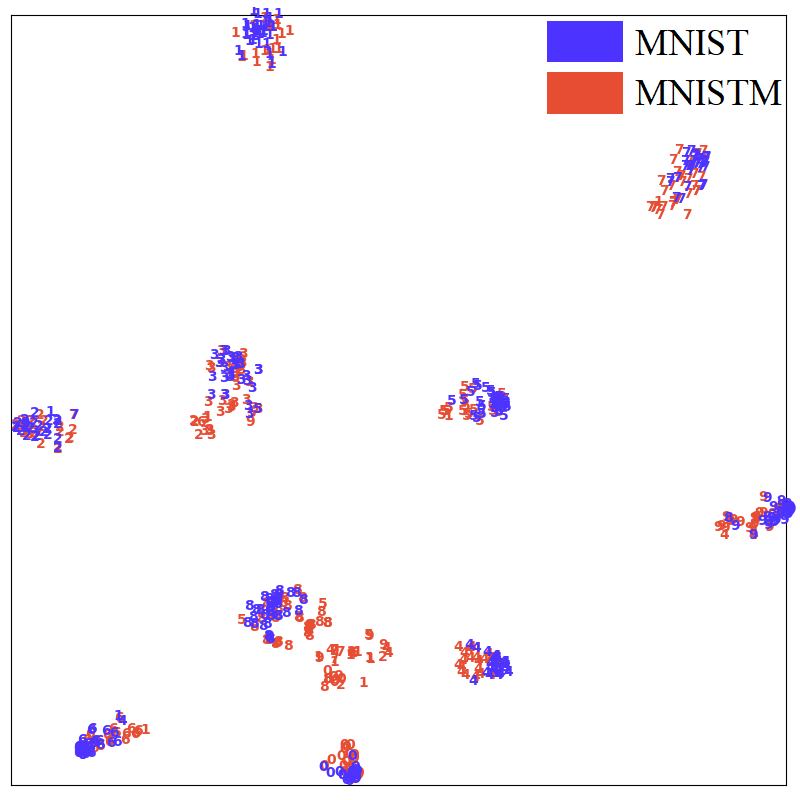}}
\hspace{-3mm}
\subfigure[CoSCA M$\rightarrow$MM]{\label{fig:tsne_mnistm_ours}
\includegraphics[width=0.25\textwidth]{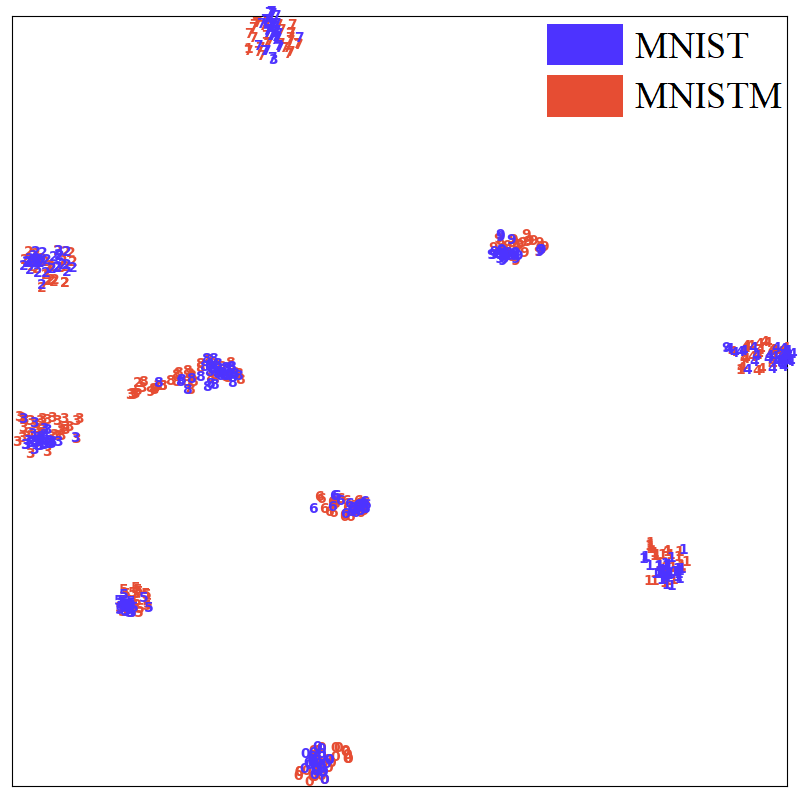}}
\vspace{3mm}
\caption{t-SNE embedding of the features $G(x)$ for MNIST (M) $\rightarrow$USPS (U), and MNIST (M) $\rightarrow$MNISTM (MM). Color indicates domain, and the digit number is the label. The ideal situation is to mix the two colors with the same label, representing domain-invariant features.}
\label{fig:tsne_supp}
\vspace{-2mm}
\end{figure}

\noindent We also include t-SNE plots for other benchmark datasets in the following. Figure~\ref{fig:tsne_supp} compares MCD and CoSCA on MNIST$\rightarrow$USPS and MNIST$\rightarrow$MNISTM, showing that CoSCA provides improvement over MCD.


\setlength{\tabcolsep}{2pt}
\begin{table*}[!t]
\vspace{-4mm}
  \centering
  \begin{tabular}{l|cccccccccccc|c}
  \toprule
    Model & plane & bcycl & bus & car & horse & knife & mcycl & person & plant & sktbrd & train & truck & mean\\
    \midrule \midrule
    CAN [31] & 94.5 & 76.3 & 82.2 & \textbf{71.1} & 94.3 & 86.2 & 88.3 & 81.2 & 91.4 & 89.2 & 87.3 & 50.3 & 82.7\\
    SEDA [38] & 95.9 & 87.4 & 85.2 & 58.6 & \textbf{96.2} & \textbf{95.6} & 90.6 & 80.0 & 94.8 & 90.8 & \textbf{88.4} & \textbf{47.9} & 84.3\\
    CoSCA & \textbf{96.3} & \textbf{87.9} & \textbf{86.1} & 69.8 & 95.9 & 93.7 & \textbf{91.2} & \textbf{84.1} & \textbf{95.1} & \textbf{90.9} & 86.3 & 45.8 & \textbf{85.3} \\
  \bottomrule
  \end{tabular}
  \vspace{3mm}
  \caption{VisDA validation set results using a ResNet152 model.} 
  \label{tab:visda_result_2}
\end{table*}

